\documentclass[letterpaper, 10 pt, conference]{ieeeconf}

\IEEEoverridecommandlockouts
\usepackage{times}

\usepackage[hyphens]{url}  %
\usepackage{graphicx} %
\usepackage[style=ieee,backend=bibtex,url=false,doi=false,natbib=true,mincitenames=1,dashed=false,isbn=false]{biblatex}
\AtEveryBibitem{
    \ifentrytype{inproceedings}{
         \clearlist{address}
         \clearlist{publisher}
         \clearname{editor}
         \clearlist{organization}
         \clearfield{url}  
         \clearfield{doi}  
         \clearfield{pages}  
         \clearlist{location}
         \clearlist{month}
     }{}
     \ifentrytype{article}{
         \clearlist{address}
         \clearlist{publisher}
         \clearname{editor}
         \clearlist{organization}
         \clearfield{url}  
         \clearfield{doi}  
         \clearlist{location}
     }{}
 }

\usepackage{multicol}
\usepackage[bookmarks=true,hidelinks]{hyperref}

\graphicspath{{../}{./}{./figures/}}
\usepackage[utf8]{inputenc}

\usepackage[font=small,labelfont=bf]{caption}

\usepackage{graphicx}
\usepackage{xcolor}
\usepackage{amsmath}
\usepackage{amsfonts}
\usepackage{amssymb}
\usepackage{mathtools}
\usepackage{todonotes}
\let\labelindent\relax
\usepackage{enumitem}
\usepackage{comment}
\usepackage{booktabs}
\usepackage{multirow}
\usepackage[per-mode=symbol]{siunitx}
\DeclareSIUnit\minute{min}

\usepackage[noend]{algorithm,algorithmic}
\floatname{algorithm}{Algorithm}

\usepackage{tikz}
\usetikzlibrary{shapes.geometric, arrows}
\usepackage{tikzpagenodes}

\usepackage{parskip}
\usepackage{fancyhdr}

\usepackage{subcaption}

\usepackage{textcomp}
\usepackage{gensymb}

\usepackage[normalem]{ulem} % for strikethrough /sout

\def\BibTeX{{\rm B\kern-.05em{\sc i\kern-.025em b}\kern-.08em
    T\kern-.1667em\lower.7ex\hbox{E}\kern-.125emX}}

\newcommand{\ra}[1]{\renewcommand{\arraystretch}{#1}}

\newcommand{\ncomment}[1]{}

\newcommand{\sign}{\text{sign}}

\newsavebox\CBox
\def\mathBF#1{\sbox\CBox{$#1$}\resizebox{\wd\CBox}{\ht\CBox}{$\mathbf{#1}$}}

\usepackage{flushend}

\usepackage{cleveref}

\DeclareUnicodeCharacter{0301}{*************************************}

\begin{document}

\newcommand{\norm}[1]{\left\lVert#1\right\rVert}
% \title{Human-Robot Collaboration with Path Preferences}

\title{Incorporating Human Path Preferences in \\ Robot Navigation with Minimal Interventions\\
\thanks{
The work is partially supported by the Jet Propulsion Laboratory, California Institute of Technology, under a contract with the National Aeronautics and Space Administration (80NM0018D0004), and funded through JPL’s Strategic University Research Partnerships (SURP) program.}
}

\author{Oriana Peltzer$^{1}$, Dylan M. Asmar$^{2}$, Mac Schwager$^{2}$, Mykel J. Kochenderfer$^{2}$% <-this % stops a space
\thanks{Department of Mechanical Engineering, Stanford University (e-mail: peltzer@stanford.edu).}  %
\thanks{Department of Aeronautics and Astronautics, Stanford University (e-mail: \{asmar, schwager, mykel\}\!@stanford.edu).}%
}

\maketitle
\thispagestyle{empty}
\pagestyle{empty}

\begin{abstract}
Robots that can effectively understand human intentions from actions are crucial for successful human-robot collaboration. In this work, we address the challenge of a robot navigating towards an unknown goal while also accounting for a human’s preference for a particular path in the presence of obstacles. This problem is particularly challenging when both the goal and path preference are unknown a priori. To overcome this challenge, we propose a method for encoding and inferring path preference online using a partitioning of the space into polytopes. Our approach enables joint inference over the goal and path preference using a stochastic observation model for the human. We evaluate our method on an unknown-goal navigation problem with sparse human interventions, and find that it outperforms baseline approaches as the human’s inputs become increasingly sparse. We find that the time required to update the robot’s belief does not increase with the complexity of the environment, which makes our method suitable for online applications. 
\end{abstract}

% \IEEEpeerreviewmaketitle

% NOTATIONS
\newcommand{\location}{s} %
\newcommand{\goal}{g} %
\newcommand{\goalset}{\Omega_g}
\newcommand{\preference}{\theta}
\newcommand{\preferenceset}{\Theta}
\newcommand{\action}{a}
\newcommand{\actionspace}{\mathcal{A}}
\newcommand{\observation}{o}
\newcommand{\projectedlocation}{o}
\newcommand{\rewardfunction}{R}

\section{Introduction}

Collaboration between humans and robots has become increasingly important and one key aspect of this collaboration is the ability for robots to adapt to human decisions. In many scenarios, such as a robot navigating through a busy room to deliver an item, it is important for the robot to take into account human preferences. For instance, humans may prefer a specific path that would allow their colleagues to notice the item being delivered, but this preference may change dynamically based on various factors such as changes in the environment or unforeseen circumstances. While some preferences can be incorporated into the path-planning process, accommodating dynamic user preferences in real-time remains challenging. In this paper, we propose a way to enable robots to adapt to human preferences dynamically by leveraging real-time feedback to inform decision-making.

In this work, we tackle the problem of robot navigation in which the robot cannot observe the goal or the preferred path to the goal, but must make navigation decisions that are influenced by humans through recommended actions. Prior work has explored how to adapt to a human's preference through feedback, but such approaches often require a high level of intervention, which can be time-consuming and impractical in real-world scenarios. To optimize the use of human input and quickly infer the human's preference, we propose an approach that leverages probabilistic representations of human preference and incorporates real-time feedback.

Previous research by \citet{bajcsy2017learning} considered an online adaptation problem in a manipulation task, where the person can apply forces to the robot to indicate their preferences. By allowing the robot to continue its task while taking into account a probabilistic representation of human preference, their approach does not require frequent inputs. Building on this idea, we adopt a similar approach to adapt to a human's preference in the context of a robot autonomously navigating through a known environment, such as a cluttered office space. Specifically, we focus on allowing the human to influence the robot's trajectory with respect to obstacles, by providing guidance on preferred routes or paths, while the robot continues to execute its task.

\begin{figure}
    \centering
    \includegraphics[width=0.76\columnwidth]{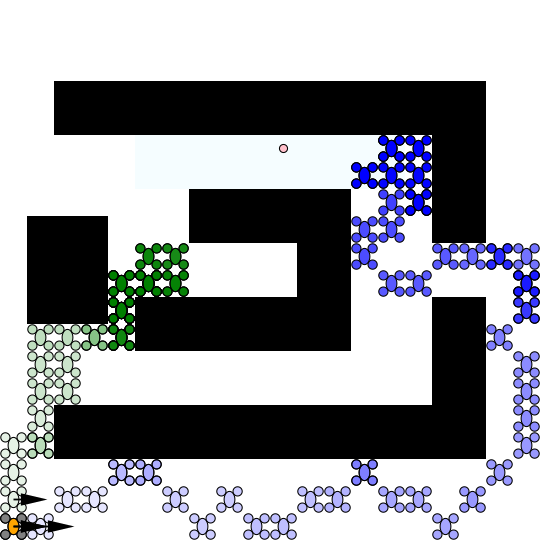}
    \caption{An autonomous robot navigates in a simulated classroom towards a goal location (pink circle). At the start of its mission, it receives direction indications (arrows) from a human that indicate which path it should take to get to the goal. In this scenario, the human wants the robot to go around the desks on the right side of the classroom. A robot that does not reason over path preferences (green) will take the shortest path to the goal regardless of the human's input. Our method (blue) infers the human's path preference from these indications and adapts to their recommendations.}
    \label{fig:Map3}
\end{figure}

Paths can be represented using homotopy classes \cite{bhattacharya2012topological}. However, homotopies can pose computational challenges when used to encode and infer human preferences. When the robot maintains a belief over homotopy classes, the inference problem can become exponentially complex with the number of obstacles in the space. Additionally, when the goal is unknown, the number of variables increases with the number of candidate destinations. This complexity can render the decision-making problem intractable. 

Our solution is to encode path preference based on a partitioning of the environment into polytopes \cite{vincent2021reachable}. This representation allows path preferences to be expressed as sets of preferred transitions between adjacent polytopes. Paths belonging to different homotopy classes correspond to different sequences of transitions. By leveraging conditional independence assumptions, we can make the Bayesian inference problem tractable. These assumptions exploit the fact that human actions provide information about the path in a piece-wise manner. For example, indicating a preference for navigating around a particular obstacle only provides information about the local area and not the entire path. Finally, after updating its belief representation over the human's preference, the robot can adapt to indications by replanning online.

Our contributions are as follows.
\begin{itemize}
    \item We formulate the human-robot collaboration problem as a Partially Observable Markov Decision Process (POMDP) where both the goal of the task and the human's path preference are unknown random variables.
    \item We propose an encoding of a human's path preference using a partitioning of the environment into polytopes, along with conditional independence assumptions that make the Bayesian inference problem tractable to infer the task goal and path preference online.
    \item Through simulations in two environments of different sizes and complexity, we show that our method is effective for solving problems where the robot must reach a goal that is unknown a-priori while simultaneously adapting to a human's indications. Our method shows higher success rates compared to baseline approaches when the human inputs are sparse. 
\end{itemize}

Our approach enables a robot to make effective navigation decisions in collaboration with a human, even when the goal and path preference are not known in advance, and with minimal human input.

\section{Related Work}

\begin{figure}
\centering
\begin{tikzpicture}[->,>=stealth',auto,node distance=1.6cm,
  thick,baseline={(0,0)}]
  
  \tikzstyle{dots} = [circle,text centered]
  \tikzstyle{state} = [circle,draw,fill=black!5, minimum size=1.0cm]
  \tikzstyle{observation} = [circle,draw,fill=blue!5, minimum size=1.0cm]
  \tikzstyle{belief} = [circle,draw,fill=orange!5, minimum size=1.0cm]

  \node[dots] (1) [below=5mm] {\footnotesize\dots};
  \node[state] (2) [right of=1, left=-1mm] {\footnotesize$\location_{t-1}$};
  \node[state] (3) [right of=2] {\footnotesize$\location_t$};
  \node[state] (4) [right of=3] {\footnotesize$\location_{t+1}$};
  \node[dots] (5) [right of=4, left=1mm] {\footnotesize\dots};
  
  \node[observation] (6) [below of=2] {\footnotesize$o_{t-1}$};
  \node[observation] (7) [below of=3] {\footnotesize$o_t$};
  \node[observation] (8) [below of=4] {\footnotesize$o_{t+1}$};
  
%   \node[belief] (goal) [above of=3] {$g$};
  \node[belief] (goal) [below of=7] {\footnotesize$g, \theta$};

  \path[every node/.style={font=\sffamily\small}]
    (1) edge node [right] {} (2)
    (2) edge node [right] {} (3)
    (3) edge node [right] {} (4)
    (4) edge node [right] {} (5)
    (2) edge node [below] {} (6)
    (3) edge node [below] {} (7)
    (4) edge node [below] {} (8)
    % (goal) edge node [left] {} (3)
    % (goal) edge node [right] {} (4)
    % (goal) edge node [left] {} (2);
    (goal) edge node [left] {} (6)
    (goal) edge node [right] {} (7)
    (goal) edge node [left] {} (8);
    %(4) edge[bend right] node [left] {} (1);
    % (goal) edge [bend right] node [left] (3)

    % \draw [dotted, draw=gray, ->] (goal) -- (2);
    % \draw [dotted, draw=gray, ->] (goal) to [bend right] (3);
    % \draw [dotted, draw=gray, ->] (goal) -- (4);

\end{tikzpicture}
\caption{We model the intent inference problem with the above diagram. At each step in time, the robot receives an observation $o_t$ from the human conditioned on its current location $\location_t$, the intended goal $\goal$, and the human's path preference $\preference$. The robot updates its belief over $\goal$ and $\preference$ and transitions to a next location $\location_{t+1}$. %Variants of this problem may include a direct effect of the observation on the robot's next state (for example, the human may apply a force on the robot that changes its trajectory).
}
\label{fig:BII_diagram_policy_obs}
\end{figure}
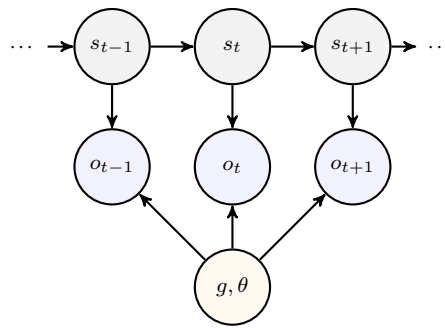

% \begin{figure}
% \label{BII_diagram_complete}
% \begin{tikzpicture}[->,>=stealth',auto,node distance=2.5cm,
%   thick]
  
%   \tikzstyle{dots} = [circle,text centered]
%   \tikzstyle{state} = [circle,draw,fill=black!5, minimum size=1.2cm]
%   \tikzstyle{belief} = [circle,draw,fill=orange!5, minimum size=1.2cm]

%   \node[dots] (1) {\dots};
%   \node[state] (2) [right of=1] {$S_{t-1}$};
%   \node[state] (3) [right of=2] {$S_t$};
%   \node[state] (4) [right of=3] {$S_{t+1}$};
%   \node[dots] (5) [right of=4] {\dots};
  
%   \node[belief] (goal) [above of=3] {$g$, $\pi$};

%   \path[every node/.style={font=\sffamily\small}]
%     (1) edge node [right] {} (2)
%     (2) edge node [right] {} (3)
%     (3) edge node [right] {} (4)
%     (4) edge node [right] {} (5)
%     (goal) edge node [left] {} (3)
%     (goal) edge node [right] {} (4)
%     (goal) edge node [left] {} (2);
%     %(4) edge[bend right] node [left] {} (1);
% \end{tikzpicture}
% \caption{Intent inference model for trajectory prediction with a path preference. 
% % Version 1: The belief in the human's preferred sequence of hyperplanes is modeled as a discrete probability distribution over the set of all possible sequences for the solution. 
% % Version 2: 
% The belief in the human's planning policy is modeled as a discrete probability distribution over the set of all possible sequences for the solution for each robot location.}
% \end{figure}

In recent years, there has been a growing interest in shared autonomy and interactive systems, where humans and robots work together to accomplish tasks. Several approaches have been proposed to address the challenge of enabling effective collaboration between human and robot agents while still achieving high task performance. \citet{losey2020controlling, jeon2020shared} propose a framework where a human operator is given control of a task-relevant latent action space while an autonomous system handles the rest. \citet{dragan2013policy} present a formalism for arbitrating between a user's input and a robot's policy when both human and robot share control of the same action space. \citet{Cognetti2020PerceptionAwareHN} provide a method for real-time modifications of a path, while \citet{hagenow2021corrective} present a method that allows an outside agent to modify key robot state variables and blends the changes with the original control. However, a common challenge of these approaches is the high level of intervention required from humans.

% -------------------------------------------
% %Figure placement

\begin{figure*}[t!]
     \centering
     \subcaptionbox{Hyperplane arrangement of a two-dimensional space containing two obstacles (colored in gray). The robot is located inside the pink polytope, surrounded by three adjacent obstacle-free polytopes. Each hyperplane on the boundary of the robot's polytope corresponds to one of the non-redundant constraints in \cref{eq:polytope_h_representation_essential}.\label{fig:HA}}%
     [0.28\textwidth]{\includegraphics[width=0.28\textwidth]{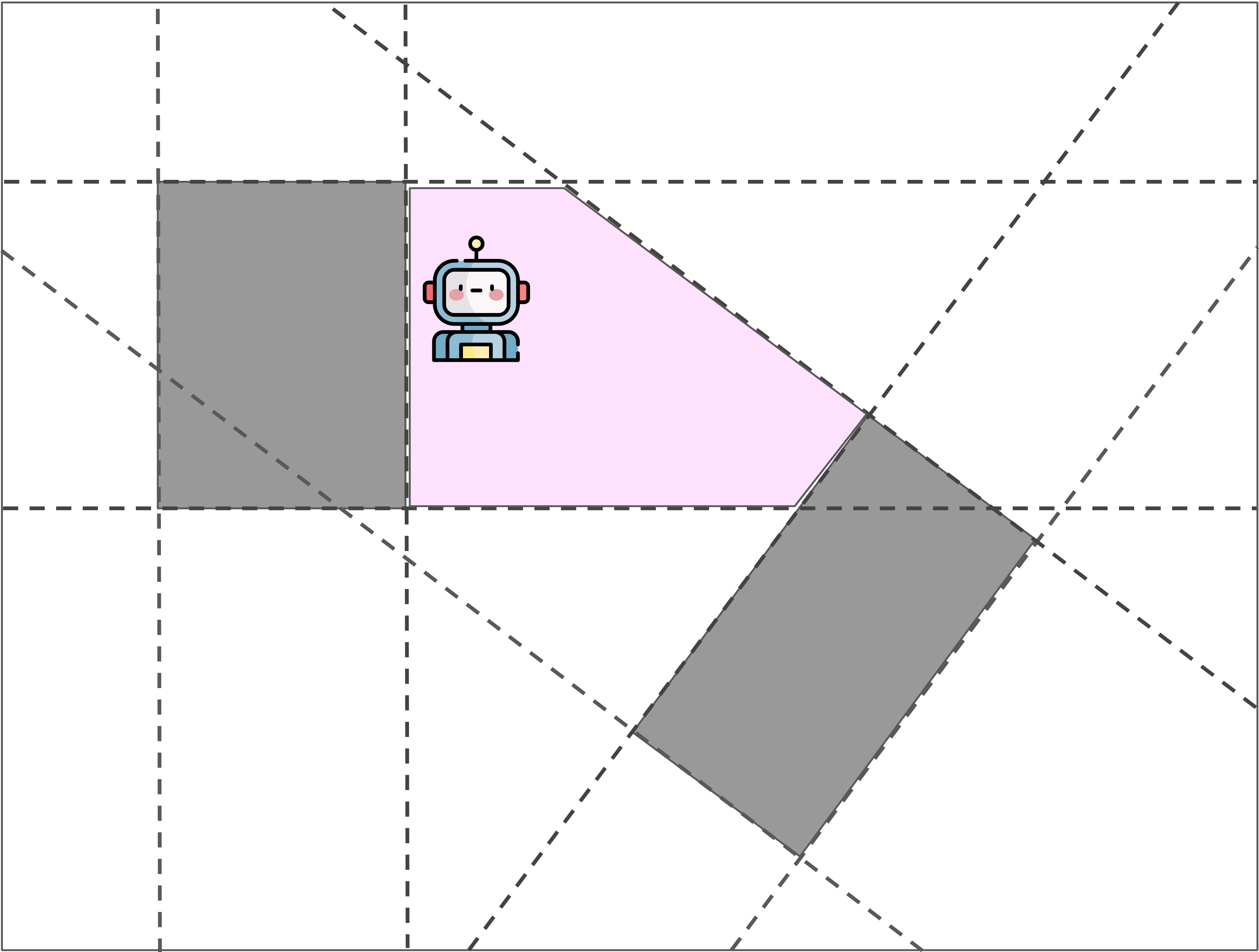}} 
     \hfill
     \subcaptionbox{Graph derived from the hyperplane arrangement. The nodes on the graph designate polytopes, and edges designate transitions to adjacent polytopes. To estimate the human's preference, the robot updates a posterior over the goal and over which of the graph transitions $\phi_1$, $\phi_2$ and $\phi_3$ is preferred by the human.\label{fig:HA_graph}}%
     [0.28\textwidth]{\includegraphics[width=0.28\textwidth]{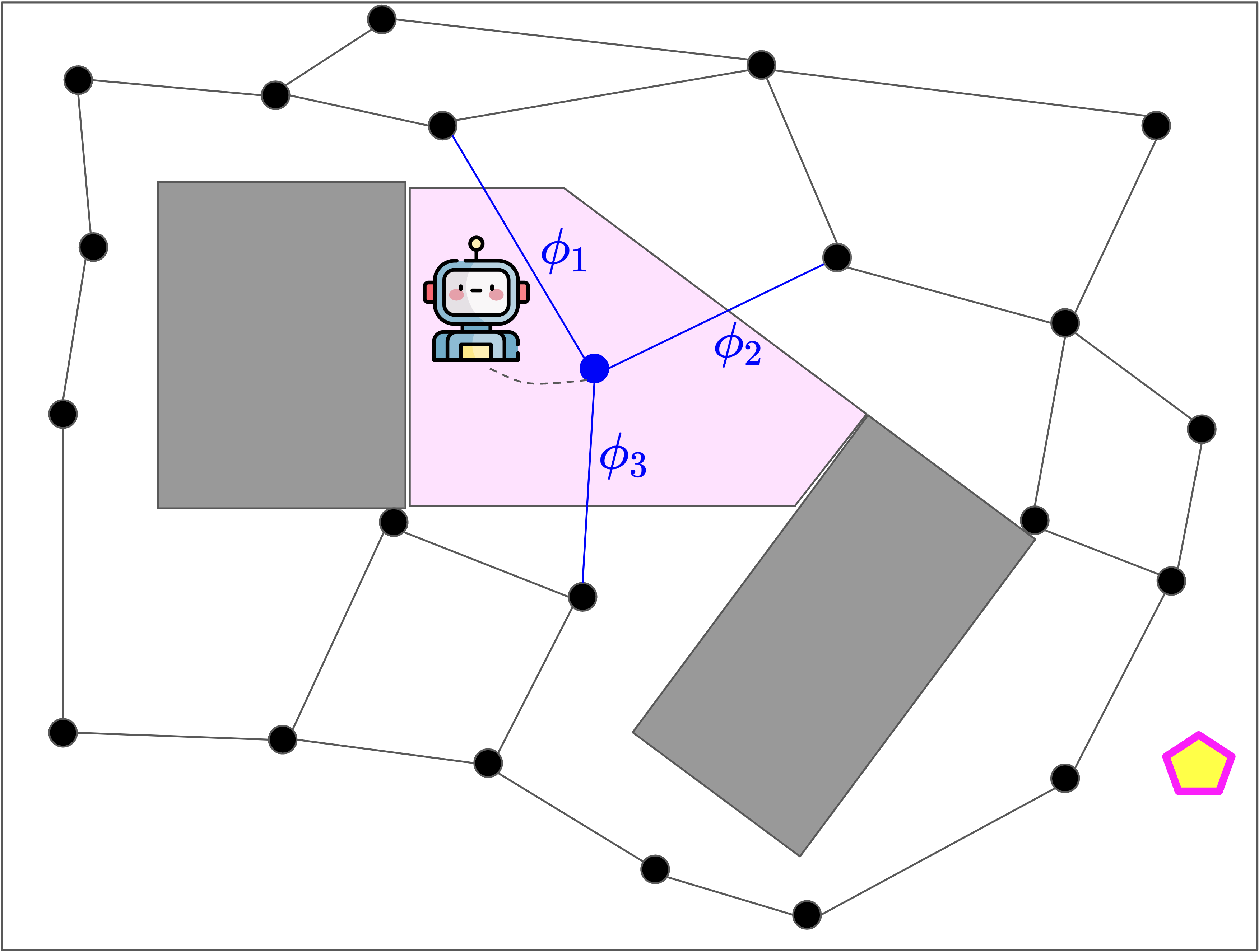}} 
     \hfill
     \subcaptionbox{Example preference defined over the graph. The location of the goal is indicated in yellow in the lower right polytope. For each node, the outgoing pink arrow designates the edge on the graph corresponding to the preferred transition between polytopes.\label{fig:HA_pref}}%
     [0.28\textwidth]{\includegraphics[width=0.28\textwidth]{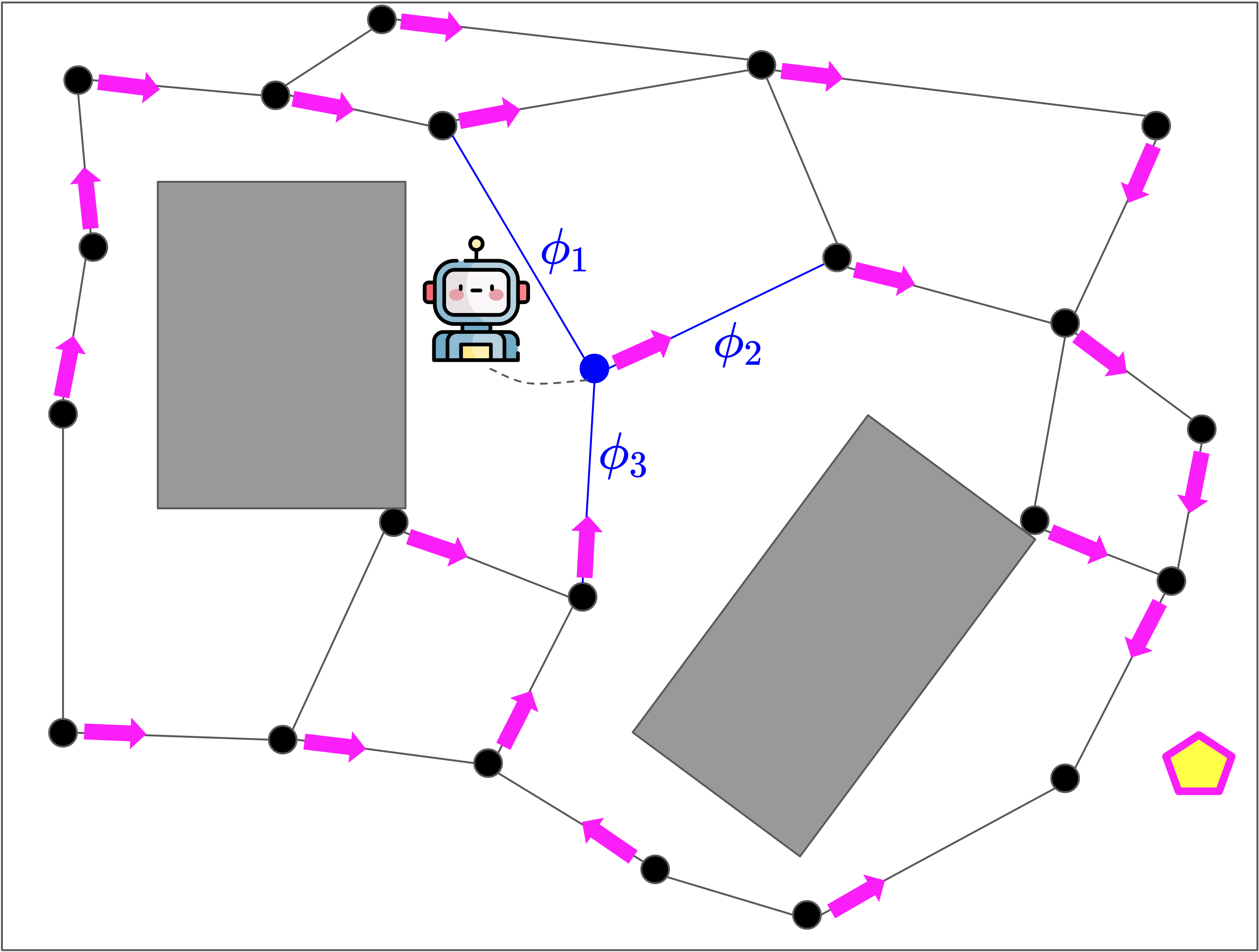}} 
    \caption{Using the hyperplanes composing the H-representation of each obstacle, we construct a hyperplane arrangement of the obstacle-free space (a). We define the human's preference for the robot's one step action choices as the posterior distribution (given all human input up to that point) over transitions from the current to the neighboring polytopes, i.e. edges on the graph. Each time the robot transitions to a new polytope, the set of neighbor polytopes and the distribution over human preferences are updated.}
        \label{fig:hyperplane_arrangement}
\end{figure*}

\citet{best2015bayesian} propose a method for predicting an agent's intended trajectory from observations. Rather than maintaining a belief over the agent's future path, they infer the agent's intended goal among a set of candidate locations at the boundary of the space. This approach provides information on where the agent is heading and generates a distribution of candidate future trajectories for the agent. Inferring the goal of the task among a discrete set of candidates is also relevant to the area of shared autonomy. \citet{javdani2015shared} propose a formalism for shared control of a robotic arm, where the robot must assist the human in picking up an object but needs to infer which object the human has chosen from joystick inputs.

Planning with homotopy class constraints is useful in problems where the robot's requirements are given with respect to obstacles, and \citet{yi2016homotopy} consider topological constraints provided by human operators. \citet{Bhattacharya_2010} propose an efficient algorithm for solving path-planning problems under homotopic constraints. However, the number of homotopy classes for a given problem can be infinite, and as the robot changes location and updates its representation of the world, carrying out inference over homotopy classes in a dynamic environment requires re-computing the set of homotopies at every iteration, making the belief update challenging.

Prior work has addressed the challenge of shared autonomy by considering how robots can infer a human's intended goal, or how they can infer the preferred path to a goal. However, we argue that inferring the goal and the path as separate problems can lead to over-confidence in incorrect beliefs about the user's preferences. To illustrate this point, consider the following scenario: a robot and a human are collaborating to move an object from one end of a room to another, but there is an obstacle in the way. The human would like the robot to take a path around the obstacle on the left, even though the goal is on the right. If the robot only infers the goal from the human's inputs, it may incorrectly assume that the goal is on the right, and become over-confident in this belief. On the other hand, if the robot only infers the preferred path, it may mistakenly assume that the goal is on the left, leading to a failure in completing the task.

To overcome these challenges, our work proposes a joint inference approach that considers both the human's intended goal and their preferred path to that goal. Specifically, we model the human's preference over different homotopy classes and leverage a conditional independence assumption to provide a tractable solution. In our approach, we assume that the human's inputs are noisily rational conditioned on both the goal and the preference. By jointly inferring the goal and path preference, we can avoid over-confidence in incorrect beliefs about the user's preferences, leading to improved system performance.

\section{Problem Statement}

We consider the problem of robot navigation in a known environment to an unknown destination, where a human can intervene and provide a heading direction to the robot using a joystick or force cues. The human also has a preference on which path the robot should take with respect to obstacles, and our objective is for the robot to understand the human's intentions and execute the task with minimal interventions.

Let $\goal$ be a discrete random variable denoting the goal of the task, belonging to a set of candidates $\goalset$, and let $\preference$ be a discrete-valued random variable representing the human's path preference, belonging to a set of possible preferences $\preferenceset$. The physical location of the robot at time index $t$ is denoted by $\location_t \in \mathbb{R}^2$, and the robot's action at time index $t$, belonging to some action space $\actionspace$, is denoted by $\action_t$. The transition model $T(\location_{t+1} \mid \location_t,\action_t)$ is deterministic, meaning the robot has full control over its future location. At any time step, the human may provide an observation to the robot. When the human intervenes, the robot receives a direction (heading angle) that can be mapped to a future location in space. More specifically, we map the direction to an intended location, which is the resulting robot location after advancing in the indicated direction for one time step. For simplicity, we consider that the robot directly makes an observation $\observation_t$ of the location indicated by the human.

We assume that the robot has a stochastic observation model for the human $P(\observation_t \mid \location_t, \goal, \preference)$ that is conditioned on both the goal of the task $\goal$ and the human's preferred path $\preference$. We further assume that having chosen a goal and path preference, the human takes actions to noisily minimize a cost function $C_{\goal,\preference}$ that measures the cost of moving from the robot's current location to the goal along the preferred path. For example, $C_{\goal,\preference}(\location_t, \projectedlocation_t)$ can be the length of the shortest path from location $\location_t$ to the goal $\goal$ after taking a first step to $\projectedlocation_t$, and constrained by path preference $\preference$.

We use $C_{\goal,\preference}$ to induce a probability distribution over observations, given by:
\begin{equation}
\label{eq:humanmodel}
P(\observation_t \mid \location_t, \goal, \preference) \propto e^{-\gamma_h \cdot C_{\goal,\preference}(\location_t, \projectedlocation_t)},
\end{equation}%
where $\gamma_h$ is a hyperparameter that designates the rationality coefficient.%, and $C^{\star}_{\goal,\preference}(\location_t)$ is the minimum cost achievable by the robot from location $\location_t$ to the goal $\goal$ along the preferred path $\preference$.

This model assumes the human will pick the lowest cost action with the highest probability and the likelihood of an action decreases exponentially with the increase in cost \cite{ziebart2008maximum}. Our inclusion of the path preference $\preference$ sets our approach apart from \cite{best2015bayesian}. The model is shown in \cref{fig:BII_diagram_policy_obs} represented as a Bayesian Network.

\subsection{Inference}

At each time step where the human provides an observation, the posterior $P(g,\theta)$ is given through the Bayesian update
\begin{align}
    &P(g,\theta \mid \projectedlocation_{1:t}, \location_{1:t})  \; \propto \; \nonumber \\
    &\qquad\qquad\;\; P(\projectedlocation_{t} \mid \location_t, \goal, \preference)  P(\goal,\preference \mid \projectedlocation_{1:t-1}, \location_{1:t}). \label{eq:inference}
\end{align}
We note that the number of Bayesian updates required at each iteration to update the belief is equal to the cardinality of $\goalset \times \preferenceset$. In addition, each Bayesian update involves computing $C_{\goal, \preference}(\, . \, , \, . \,)$ in \cref{eq:humanmodel}, which involves solving an optimization problem (such as a shortest path problem).
%If policy $\theta$ represents a discrete distribution over all possible paths to the goal starting from any current state, the above update is intractable. 
In \cref{sec:path_preference}, we propose a specific encoding of preference $\preference$ for resolving \cref{eq:inference}, while ensuring the number of computations of the cost $C_{\goal, \preference}(. , .)$ per update does not grow exponentially with the number of obstacles.

\subsection{Decision Making}
\label{subsec:pomdp}

We consider a navigation problem where the robot receives reward according to the model $R(\location_t, \goal, \preference, \action_t)$.
We wish to find the optimal policy $\pi$ that maximizes the expected discounted sum of future rewards, with discount factor $\gamma$.
The above problem is a Partially Observable Markov Decision Process (POMDP) \cite{kochenderfer2015decision}.%described by the tuple $(\mathcal{S}, \actionspace, \mathcal{O}, T, \mathcal{R}, \gamma)$ \cite{kochenderfer2015decision}.

\section{Path Preference}
\label{sec:path_preference}

In this section, we propose an encoding of human's path preference $\preference$ for computing the posterior in \cref{eq:inference}. Deviating from the concept of homotopy classes, we define the preference according to a partitioning of the environment into polytopes, as shown in \cref{fig:hyperplane_arrangement}, creating a hyperplane arrangement of the space. Hyperplane arrangements have been used by \citet{vincent2021reachable} in the context of Neural Network verification.
In our setting, we leverage this representation to define path preferences as preferred transitions between adjacent regions of the space.

\subsection{Hyperplane Arrangement}

\begin{figure}
\centering
\begin{tikzpicture}[->,>=stealth',auto,node distance=1.8cm,baseline={(0,0)},
  thick]
  
  \tikzstyle{dots} = [circle,text centered]
  \tikzstyle{state} = [circle,draw,fill=black!5, minimum size=1.0cm]
  \tikzstyle{belief} = [circle,draw,fill=orange!5, minimum size=1.0cm]
  \tikzstyle{homotopy} = [circle,draw,fill=red!50!orange!50!white, minimum size=1.0cm]
  \tikzstyle{observation} = [circle,draw,fill=blue!5, minimum size=1.0cm]

  % \node[dots] (1) {\dots};
  % \node[state] (2) [right of=1] {$S_{t-1}$};
  % \node[state] (3) [right of=2] {$S_t$};
  % \node[state] (4) [right of=3] {$S_{t+1}$};
  % \node[dots] (5) [right of=4] {\dots};

  \node[dots] (1) [below=5mm] {\footnotesize\dots};
  \node[state] (2) [right of=1, left=-1mm] {\footnotesize$\location_{t-1}$};
  \node[state] (3) [right of=2] {\footnotesize$\location_t$};
  \node[state] (4) [right of=3] {\footnotesize$\location_{t+1}$};
  \node[dots] (5) [right of=4, left=1mm] {\footnotesize\dots};

  \node[observation] (6) [below of=2] {\footnotesize$o_{t-1}$};
  \node[observation] (7) [below of=3] {\footnotesize$o_t$};
  \node[observation] (8) [below of=4] {\footnotesize$o_{t+1}$};
  
  \node[homotopy] (h1) [below of=6] {\footnotesize$p_i$};
  \node[homotopy] (h2) [below right of=7, right=-9mm] {\footnotesize$p_j$};
  
  \node[belief] (goal) [below of=7, below=5mm] {\footnotesize$g$};

  \path[every node/.style={font=\sffamily\small}]
    (1) edge node [right] {} (2)
    (2) edge node [right] {} (3)
    (3) edge node [right] {} (4)
    (4) edge node [right] {} (5)
    (2) edge node [below] {} (6)
    (3) edge node [below] {} (7)
    (4) edge node [below] {} (8)
    (h1) edge node [left] {} (6)
    (h2) edge node [left] {} (8)
    (h2) edge node [left] {} (7)
    (goal) edge node [left] {} (h1)
    (goal) edge node [right] {} (h2)
    (goal) edge node [left] {} (6)
    (goal) edge[bend right] node [right] {} (8)
    (goal) edge node [left] {} (7);
    %(4) edge[bend right] node [left] {} (1);
\end{tikzpicture}
\caption{Intent inference model in a hyperplane arrangement of the obstacle free space. We spatially decompose the preference $\preference$ into a set of preferred neighboring polytopes per region of the space. 
Within each polytope $j$, the human preference $p_j$ is a discrete distribution over the preferred neighbor in $\mathcal{N}(j)$. We assume that for a location $\location_t$ belonging to polytope $j$, and given goal $\goal$ and preference $p_j$, the observation $\observation_t$ and any other preference $p_{i, i \neq j}$ are conditionally independent.
% Our main assumption is that this preference only has an effect on the human's (and thus system's) behavior in states belonging inside the polytope.
%For simplicity, we removed the feed-forward effect (represented as dotted lines in Figure \ref{fig:BII_diagram_policy_combined}) of the human's control over the robot's state.
}
\label{fig:BII_diagram_decomposed_homotopies}
\end{figure}
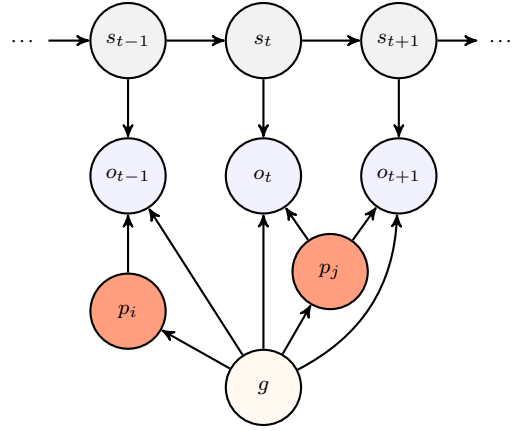

We assume a two-dimensional environment composed of $m$ polytopic obstacles, each defined by their half-space representation (H-representation)
$$\mathcal{O}_i = \lbrace x \in \mathbb{R}^2 \text{ s.t. } A_ix \leq b_i \rbrace,$$ 
where $A_i \in \mathbb{R}^{d_i \times 2}$ and $b_i \in \mathbb{R}^{d_i}$, and where $d_i$ is the number of edges (hyperplanes) composing polytope $i$. Let $n = \sum_i d_i$ be the total number of hyperplanes. 

We leverage each obstacle's H-representation to construct a hyperplane arrangement of the environment as shown in \cref{fig:hyperplane_arrangement}, i.e. a partitioning of the space into polytopes. More specifically, each location in space belongs to a polytope $j$ for which we can write an H-representation of the form
$$\mathcal{P}_j = \lbrace x \in \mathbb{R}^2 \text{ s.t. } \sign(A_ix - b_i) = \alpha^{j}_{i} \; \forall i \rbrace,$$
where $\alpha^{j}_{i} \in \lbrace -1,1 \rbrace^{d_i}$ is a vector specific to polytope $j$ and obstacle $i$ corresponding to the relative position of any point in the set with respect to each hyperplane in $\mathcal{O}_i$.

%Each polytope can be uniquely represented by a vector $\alpha^j$.% of dimension $n$, taking on values $-1$ or $1$ according to their positions with respect to each hyperplane.
Concatenating elements from each obstacle's H-representation, we can write polytope $j$'s H-representation as 
\begin{equation}
\label{eq:polytope_h_representation}
\mathcal{P}_j = \lbrace x \in \mathbb{R}^2 \text{ s.t. } \sign ( Ax - b ) = \alpha^j \rbrace,
\end{equation}
where 
$A := \begin{pmatrix} 
A_1 \\ \vdots \\ A_m
\end{pmatrix} \in \mathbb{R}^{n \times 2}$
and 
$b := \begin{pmatrix} 
b_1 \\ \vdots \\ b_m
\end{pmatrix} \in \mathbb{R}^{n}$.

Some of the constraints in \cref{eq:polytope_h_representation} (corresponding to rows of $A$, $b$ and $\alpha^j$) are redundant, i.e. the set $\mathcal{P}_j$ does not change upon their removal. We can further reduce the H-representation of a polytope to include only non-redundant constraints. By removing the rows corresponding to redundant constraints, we obtain new matrices $A_e^j, b_e^j$ and $\alpha_e^j$ such that we can write the polytope's reduced H-representation as
\begin{equation}
\label{eq:polytope_h_representation_essential}
\mathcal{P}_j = \lbrace x \in \mathbb{R}^2 \text{ s.t. } \sign ( A_e^jx - b_e^j ) = \alpha_e^j \rbrace.
\end{equation}
The non-redundant constraints correspond to edges of the polytope. In other words, as the robot continually moves in space, the first hyperplane that it will cross upon exiting the polytope will correspond to one of the polytope's non-redundant constraints. \citet{vincent2021reachable} outline an iterative method for removing redundant constraints by solving $n$ linear programs. We use this method in practice for computing $\alpha_e^j$ for each polytope.
We can now characterize each polytope by a vector $\alpha^j_e \in \lbrace -1,1 \rbrace^{n_e^j}$, where $n_e^j \leq n$ is the number of essential constraints of the polytope. The polytopes $\mathcal{P}_j$ partition the environment into a hyperplane arrangement.

\subsection{Path Preference}
In this section, we provide a definition of preference $\preference$ according to a graphical representation of the environment based on the hyperplane arrangement.
Under this representation, a path preference corresponds to a set of preferred transitions. In other words, for each polytope in the space, the human will have a preference to which neighboring polytope they wish to transition.

Let $\mathcal{G} := (\mathcal{V}, \mathcal{E})$ be an undirected graph, where vertices are obstacle-free polytopes, and edges connect two adjacent polytopes.
Each polytope is described by a unique vector $\alpha^j$ as defined in \cref{eq:polytope_h_representation}. Two polytopes are adjacent if they share non-redundant constraints (rows in \cref{eq:polytope_h_representation_essential}) corresponding to the same hyperplane (i.e. they are on opposite sides of the hyperplane).
Let $\mathcal{N}(v)$ be the set of neighbors of a vertex $v$. For each vertex, we denote $p_v$ the discrete-valued random variable describing which edge in $\mathcal{N}(v)$ the human intends to transition to.

Using this formalism, we define a path preference as the set of preferred transitions over all nodes in the graph,
\begin{equation}
    \label{eq:preference}
    \preference = \lbrace p_v \text{ s.t. } v \in \mathcal{V} \rbrace.
\end{equation}

Let $m_\preference = \prod_{v \in \mathcal{V}} \vert \mathcal{N}(v) \vert$
be the cardinality of $\preferenceset$, and $m_\goal = \vert \goalset \vert$ the number of possible goals. A priori, the number of Bayesian updates required to update the belief at every iteration should be $m_\preference \times m_\goal$.

Now, let us assume the conditional independence relationships described by the new problem diagram in \cref{fig:BII_diagram_decomposed_homotopies}. More specifically, we introduce the assumption that conditioned on a robot location $\location_t$, the goal $\goal$, and the preference for the corresponding vertex $p_v$ in the graph, the observation $o_t$ and the preference for any other vertex are conditionally independent. In other words, the observations the human provides can be defined conditioned only on the robot location, the goal, and the human's preference for its current vertex $p_v$. By introducing this assumption, each update step only requires updating the joint $(p_v, \goal)$, reducing the number of cost computations to $\vert \mathcal{N}(v) \vert \times m_\goal$. We can notice that by introducing this assumption, we removed the direct relationship between the number of polytopes in the environment and the complexity of the Bayesian update in \cref{eq:inference}.

In practice, components of $\preference$ are not mutually independent. For example, if the human preference at a vertex $v_1$ is $p_{v1}~=~(v_1,v_2)$, it is unlikely that the human will also prefer $p_{v_2}~=~(v_2, v_1)$ (turning back). We can improve our model by assuming a dependent relationship between preferences for adjacent edges, which does not significantly increase the complexity of the inference problem.

An interesting property of our encoding is that any two paths that belong to different homotopy classes will cross different sequences of polytopes, i.e. they correspond to a different sequence of edges on $\mathcal{G}$.
This can be proved by contradiction. Let us suppose that two continuous trajectories $\xi_1$ and $\xi_2$, with the same start and end points and that do not intersect any obstacle, traverse the same regions in $\mathcal{G}$ in the same order. From the construction of the hyperplane arrangement, each polytope that the paths traverse through is obstacle-free. Therefore, within each polytope, there is no obstacle in the area located in between the portions of $\xi_1$ and $\xi_2$ that belong to the region. A smooth transformation of $\xi_1$ into $\xi_2$ can be obtained by transforming each portion of $\xi_1$ belonging to the polytopes it intersects into the corresponding portion of $\xi_2$ for the same polytopes, where the extremities of the trajectory portions are connected to one another along the polytope's edges (where the same edge is crossed by both paths). Along this transformation, the paths do not intersect any obstacle, and therefore $\xi_1$ and $\xi_2$ belong to the same homotopy class.

\section{Experiments}

\begin{figure*}[t!]
     \centering
     \subcaptionbox{Map 1: Simple, $10 \times 10$, $8$ polytopes.\label{fig: map1}}%
     [0.23\textwidth]{\includegraphics[width=0.21\textwidth]{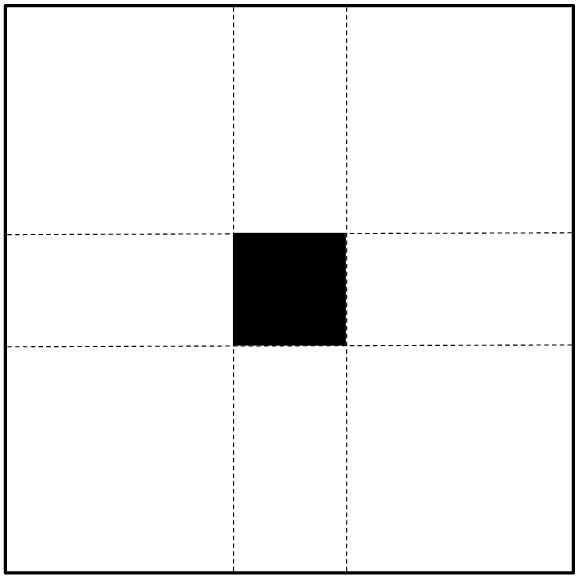}} 
     \hfill
     \subcaptionbox{Map 2: Office, $10 \times 10$, $56$ polytopes. \label{fig: map2}}%
     [0.23\textwidth]{\includegraphics[width=0.21\textwidth]{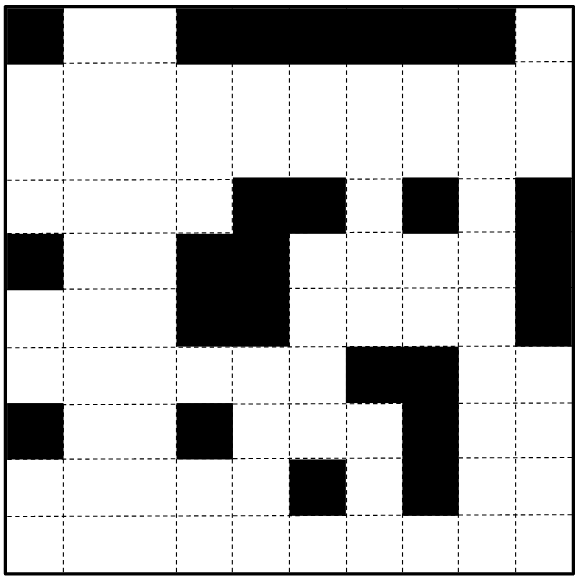}}
     \hfill
     \subcaptionbox{Map 3: Classroom, $20 \times 20$, $73$ polytopes. \label{fig: map3}}%
     [0.23\textwidth]{\includegraphics[width=0.21\textwidth]{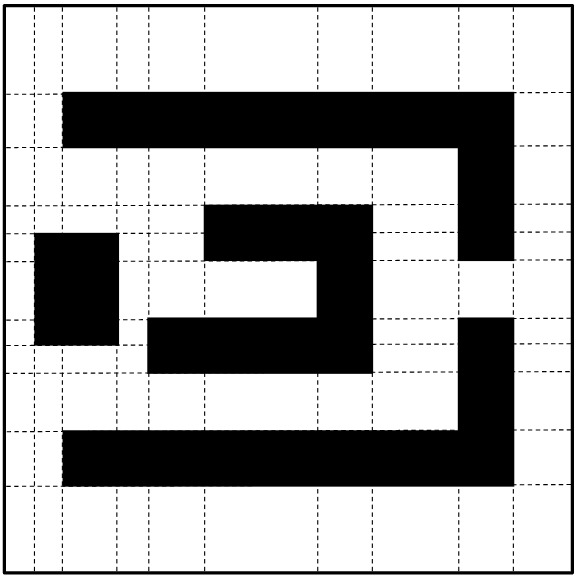}}
     \hfill
     \subcaptionbox{Sampled observations and robot's executed trajectories.}%
     [0.23\textwidth]{\frame{\includegraphics[width=0.21\textwidth]{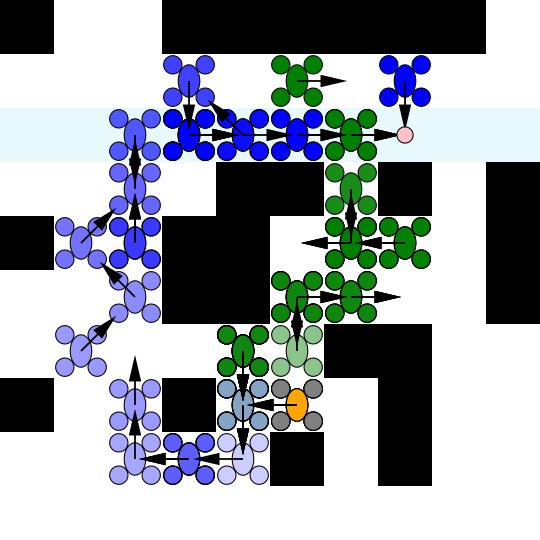}}} \label{fig: map2_samp}
    \caption{Maps used for simulating the robot navigation problem with path preferences. In (d), the heading angles observed are indicated with arrows. The goal is indicated with a pink circle, and the orange robot corresponds to the starting location. The blue robot follows a policy that accounts for path preference, while the green robot does not. The opacity of the robots increases with time.}
        \label{fig: maps}
\end{figure*}

We evaluate our model on a simulated navigation task where the robot must reach a goal that is unknown a priori while respecting the path preferences indicated by a human.

\textbf{Transition model} The robot navigates in a grid world containing obstacles. The transition model is deterministic: the robot selects an adjacent location on the grid to reach at the next time step. The robot is also allowed to take diagonal actions.
Each location $\location_t$ in the map can be mapped to a vertex $v_t \in \mathcal{G}$. Therefore, the actions leading to locations mapped to different vertices correspond to edges on the graph. We note $f(\location_t,\action_t)$ the edge crossed by taking action $\action_t$ from location $\location_t$. The robot is given a mission time limit $T_{max}$ for reaching the goal.

\textbf{Observation model} In this problem, we assume that the human selects actions to noisily minimize a cost function $\mathcal{C}_{\goal, \preference}$, where $\preference$ is defined as per \cref{eq:preference}, corresponding to the length of the shortest path to the goal constrained by the preference (where the robot is only allowed to make transitions on $\mathcal{G}$ along preferred edges). More specifically, 
\begin{equation}
\label{eq:shortestpath}
C_{\goal,\preference}(\location_t,\projectedlocation_t) = \delta(\location_t,\goal \mid \projectedlocation_t, p_{v_t}),
\end{equation}
where $\delta(\location_t,\goal \mid \projectedlocation_t, p_{v_t})$ designates the length of the shortest path from $\location_t$ to $\goal$ passing by $\projectedlocation_t$ and constrained by preference $p_{v_t}$. 
This is a slight variant of the cost function proposed by \citet{best2015bayesian}, where we add in a conditioning on the path preference. We compute costs by running the $A^{\star}$ path planning algorithm \cite{hart1968formal} on the environment maps (grid worlds with diagonal actions) and impose preference constraints by pruning invalid transitions from the search tree.

\textbf{Reward model.}
At each step in time, the robot receives a reward which is a sum of three components: a goal-specific reward 
\begin{equation}
\rewardfunction_{g}(\goal,\location_t) = 
\begin{cases}
R_g & \text{if } \location_t=\goal \\
0 & \text{otherwise,}
\end{cases}
\end{equation}
a preference-specific reward or penalty
\begin{equation}
\rewardfunction_{pref}(\location_t,\preference) = 
\begin{cases}
R_p & \text{if } f(\location_t,\action_t) = p_v \\
R_n & \text{if } f(\location_t,\action_t) \in \mathcal{E}\setminus \lbrace p_v \rbrace \\
0 & \text{otherwise,}
\end{cases}
\end{equation}
and an action cost term $\rewardfunction_{a}(\action_t) = \vert \vert \action_t \vert \vert$.
% $\rewardfunction(\location_t,\goal,\preference,\action_t) = \rewardfunction_{g}(\goal,\location_t) + \rewardfunction_{pref}(\location_t,\action_t,\preference) + \rewardfunction_{a}(\action_t)$.
The reward is discounted in time with discount factor $\gamma~=~0.95$ and we set $R_g=50$, $R_p=15.5$, and $R_n=-18$.

We compute solutions to the POMDP defined in \cref{subsec:pomdp} with the online solver POMCP \cite{silver2010monte}, and with the particularity that within the rollouts, the robot does not expect to collect human inputs. 
Each time a solution is computed, the robot takes an action and may receive an observation. If it does, it updates its belief distribution over the unknown problem variables and resolves the POMDP over a receding horizon.

\subsection{Baselines}

\begin{itemize}
    \item \textbf{Goal only.} The robot solves the POMDP %$(\mathcal{S}, \actionspace, \mathcal{O}, T, \mathcal{R}, \gamma)$ 
    while ignoring the effects of path preference. Similarly to \cite{best2015bayesian}, we assume the human is taking action to minimize a goal-dependent cost $C_{\goal}(\location_t,\projectedlocation_t) = \delta(\location_t,\goal \mid \projectedlocation_t)$, %$ - \delta^{\star}(\location_t,\goal)$
    where the conditioning on the preference is removed. We also omit the path preference's contribution to the reward $\rewardfunction_{pref}$.
    \item \textbf{Compliant.} The robot complies with the human input, but does not take an initiative. If the user stops providing information, the robot continues in the last direction indicated for $5$ time steps (conserving its momentum), then stops.
    \item \textbf{Blended.} We designed an arbitration function \cite{dragan2013policy} to decide between our proposed policy (accounting for path preferences) and the user's recommendation when the robot receives inputs. Our metric to evaluate confidence in the robot's prediction for the purpose of arbitration is the entropy of the intention distribution $H(\goal,p_i)$, where $p_i$ denotes the preferred neighbor for the current region.
    Because our representation of the world is discrete, the arbitration is given by a step function. Denoted by $U$, the action corresponding to the human's input, and $P$, the robot's prediction for the optimal action, we write the policy
    \begin{equation}
    \pi = 
    \begin{cases}
    P & \text{if } H(\goal,p_i) \leq h \\
    P & \text{if no observation is received} \\
    U & \text{otherwise,}
    \end{cases}
    \end{equation}
    where we chose $h=1.6$ as the confidence threshold.
\end{itemize}

\subsection{Results}

When evaluating the algorithm, we consider that a run is successful if the robot reached the goal within its allocated mission time $T_{max}$ and only made transitions between graph vertices corresponding to the human's preferences. We vary the time delay between human inputs, from constant guidance ($\Delta_T=1$) to only a single observation ($\Delta_T \geq T_{max})$.

\textbf{Success rates.} \Cref{table:office_results} reports the success rates for experiments conducted over six randomly sampled problem instances and $50$ runs per instance in Map 1 (\cref{fig: map1}). When the human provides inputs at every iteration, the compliant policy shows the highest success rates. However, as $\Delta_T$ increases, the compliant robot is not able to accomplish the task within the allotted time as it does not receive sufficient inputs to do so, and performance decreases compared to the autonomous baselines.
We find that in these runs, accounting for path preference consistently improves performance compared with the goal-only baseline.
Results also show that blending the user's input with the robot's policy (Path Preference + Blend) when the human provides information leads to improved performance.

\begin{figure}
     \centering
     \subcaptionbox{Map 1 problem setup and example realizations for goal-only (green) and path preference (blue) solution methods. The robot starts at the lower left corner of the environment, and the goal of the task (pink circle) is in the upper left area. The robot does not know which goal, among 10 options (shown in light blue squares), is the correct goal. The human provides noisy observations, indicated by arrows, at each iteration. The green robot selects actions according to the goal-only baseline, and the blue robot uses our proposed method to infer path preferences. The polytopes composing $\mathcal{G}$ are drawn in blue.\label{fig:setup}}%
     [0.5\textwidth]{\includegraphics[width=0.35\textwidth]{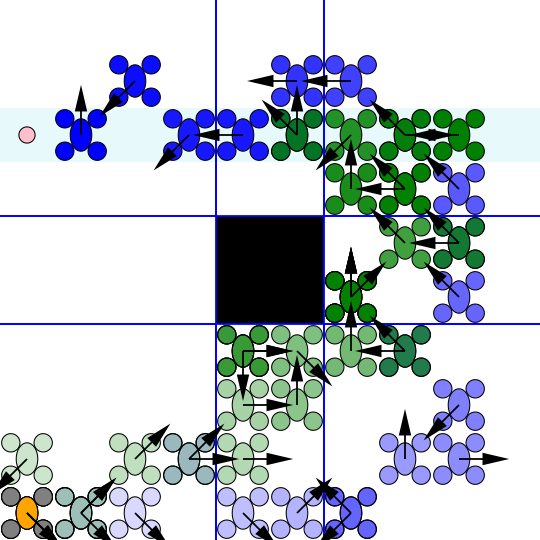}}
     \hfill
     \subcaptionbox{Probability of correct goal.\label{fig:p_correct_goal}}%
     [0.5\textwidth]{\includegraphics[width=0.5\textwidth]{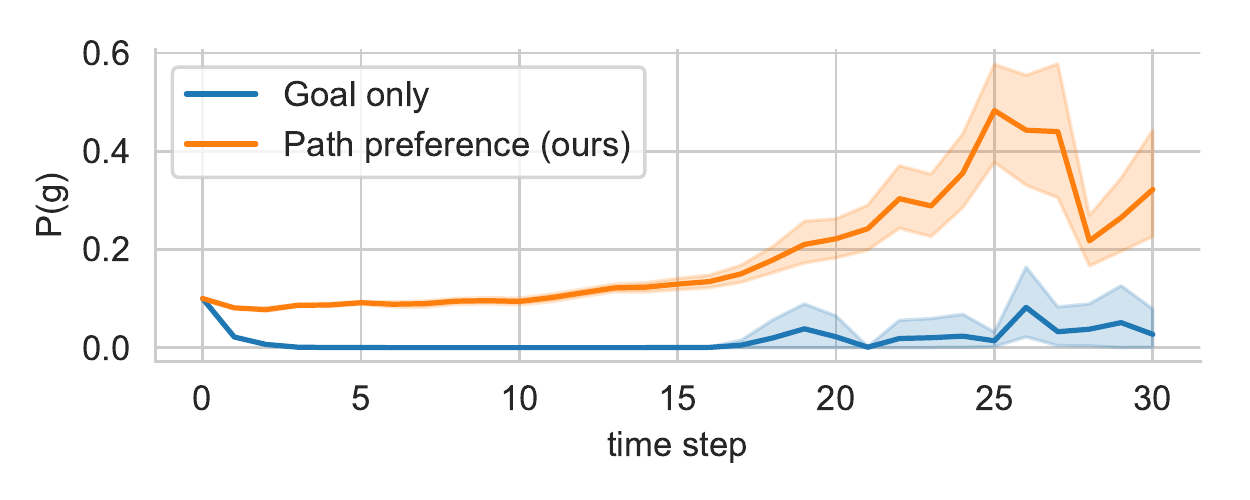}}
     \hfill
     \subcaptionbox{Entropy of goal distribution $\goal$.\label{fig:goal_dist_entropy}}%
     [0.5\textwidth]{\includegraphics[width=0.5\textwidth]{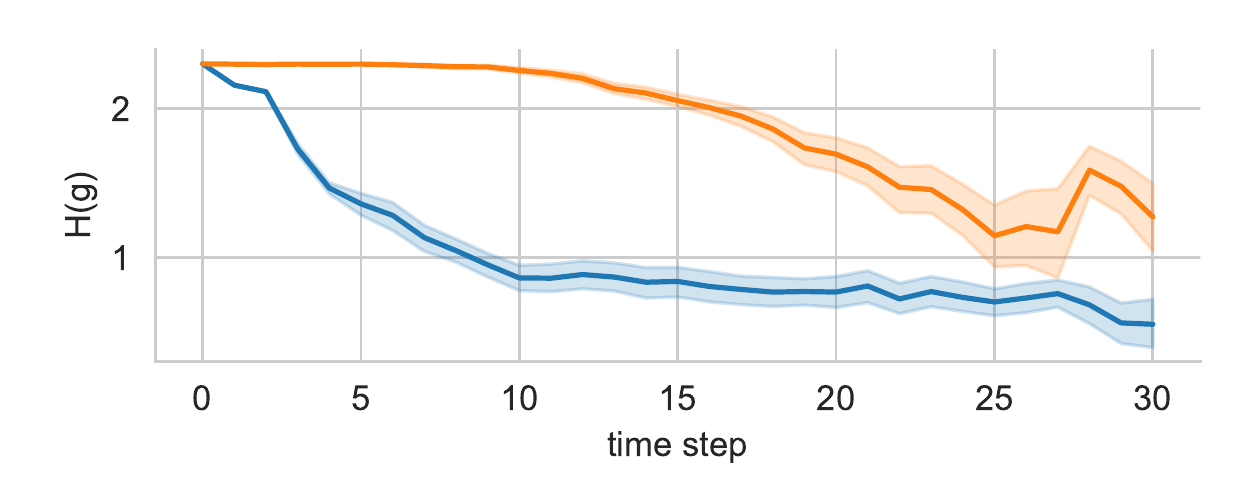}}
    \caption{Probability of the correct goal, \cref{fig:p_correct_goal}, and entropy of the goal belief distribution $P(g)$, \cref{fig:goal_dist_entropy}, for the same problem setup, \cref{fig:setup}. In this problem instance, the human's preference is to go to the goal by passing on the right side of the obstacle. Results are averaged over 50 runs and the area filled represents one standard deviation above and below the mean value. The goal-only baseline shows an over-confident prediction (shown by the strong reduction in belief entropy) that the correct goal is less likely, making it more difficult to reach the correct goal compared to a method that accounts for path preference. }
    \label{fig:overconfidence}
\end{figure}

\textbf{Belief entropy.}
\Cref{fig:overconfidence} shows a challenging problem instance where the directions the human provides do not align directly with the shortest path to the goal.
By ignoring the effects of preferences in the problem model (goal only), the robot quickly infers from observations that the upper left goal is less likely than others ($P(g)$ drops). The strong decrease in entropy shows that the robot becomes \emph{overconfident} in this prediction. Overconfidence in an incorrect goal will prevent the agent from finding the correct goal once the human's indications directly align with it, as it needs to correct for the wrong predictions, as shown in the path realization (\cref{fig:setup}). In this realization, the goal-only method (green robot) fails to search the upper left area within the allotted time. 
By accounting for path preferences in its model, the blue robot's entropy over the goal distribution decreases more steadily, allowing for it to leverage the human's latest observations and reach the goal successfully.

\begin{table}[ht]
\centering
\ra{1.3}
\begin{tabular}{@{}llrrrrr@{}}
\toprule
\multicolumn{2}{@{}l}{\multirow{2}{*}{Method}}  & \multicolumn{5}{c}{$\Delta_T$} \\
\cmidrule{3-7} 
 &&             $1$  &             $5$  &             $10$ &             $20$ &             $30$ \\
\midrule
\multicolumn{3}{@{} l }{\textbf{Map 1 (Simple) Success Rates}} \\
& Compliant               &  $\mathBF{0.67}$ &           $0.33$ &           $0.33$ &           $0.17$ &           $0.17$ \\
& Goal Only               &           $0.28$ &           $0.18$ &           $0.17$ &           $0.17$ &           $0.16$ \\
& Path Preference         &           $0.55$ &  $\mathBF{0.42}$ &           $0.41$ &           $0.34$ &           $0.30$ \\
& Path Preference + Blend &           $0.55$ &           $0.39$ &  $\mathBF{0.44}$ &  $\mathBF{0.44}$ &  $\mathBF{0.34}$ \\
\bottomrule
\end{tabular}
\caption{Success rates in the simple environment (Map 1). The results are averaged over $6$ randomly sampled problem instances  (start location, goal location, and goal possibilities), and over $50$ runs per problem instance. %The results for Map 2 are for \red{X} instances and \red{Y} runs per instance. 
$\Delta_T$ is the number of time steps separating two consecutive human inputs. The robot's mission time is $T_{max}=30$ time steps. We selected $\gamma_h=1.5$, corresponding to relatively noisy human inputs and making the problem more difficult to solve for the robot.}
\label{table:office_results}
\end{table}

\begin{table}[ht]
\centering
\ra{1.3}
\begin{tabular}{@{}llrr@{}}
\toprule
\multicolumn{2}{@{}l}{\multirow{2}{*}{Method}}  & \multicolumn{2}{c}{Computation Time (\si{\milli\second})} \\
\cmidrule{3-4} 
&& Solving & Belief update \\
\midrule
\multicolumn{2}{ @{}l }{\textbf{Map 1 (simple), $T_{max}=30$}} \\
& Goal Only                 &   $449.4 \pm \phantom{0}26.5$     &   $0.6 \pm \phantom{0}1.0$ \\
& Path Preference (ours)    &   $510.2 \pm \phantom{0}36.9$     &   $73.5 \pm \phantom{0}6.3$ \\
% RAW DATA MEAN
% goal\_only &        449.386601 &        0.564449 \\
% path\_pref &        510.231351 &       73.476734 \\
% RAW DATA STD
% goal\_only &        135.138876 &        5.058306 \\
% path\_pref &        187.999883 &       32.345589 \\
% RAW DATA 95% CI
% goal\_only &         26.487220 &        0.991428 \\
% path\_pref &         36.847977 &        6.339735 \\
\multicolumn{2}{ @{}l }{\textbf{Map 2 (office), $T_{max}=30$}} \\
& Goal Only                 &   $2984.0 \pm 149.8$     &   $0.7 \pm \phantom{0}1.0$ \\
& Path Preference (ours)    &   $2847.4 \pm 195.5$     &   $28.7 \pm 10.0$ \\
% RAW DATA MEAN
% goal\_only &       2984.065957 &        0.654573 \\
% path\_pref &       2847.440160 &       28.731385 \\
% RAW_DATA_STD
% goal\_only &        764.162018 &        4.859708 \\
% path\_pref &        997.694106 &       50.916264 \\
% RAW DATA 95% CI
% goal\_only &        149.775756 &        0.952503 \\
% path\_pref &        195.548045 &        9.979588 \\
\multicolumn{2}{ @{}l }{\textbf{Map 3 (classroom), $T_{max}=60$}} \\
& Goal Only                 &   $2305.3 \pm 101.9$              &   $0.4 \pm \phantom{0}0.8$ \\
& Path Preference (ours)    &   $2062.7 \pm \phantom{0}46.2$    &   $38.3 \pm \phantom{0}4.5$ \\
\bottomrule
\end{tabular}
\caption{Computation times for Goal Only and Path Preference methods on Map 1 (\cref{fig: map1}), Map 2 (\cref{fig: map2}), and Map 3 (\cref{fig: map3}), averaged over 100 runs with randomly sampled problem instances. The \SI{95}{\percent} confidence interval is provided with the mean. We evaluate computation time at the first iteration of each run (where the search depth takes on its highest value $T_{max}$). }
\label{table:computation_time}
\end{table}

\textbf{Computation time.} In \cref{table:computation_time} we provide the time required to solve the POMDP, and the time required to update the robot's belief as it receives new observations. We compute solutions on three maps: a simple $10\times10$ grid world with $8$ polytopes (\cref{fig: map1}), a $10\times10$ grid world with $56$ polytopes (\cref{fig: map2}), and a $20\times20$ grid world with $73$ polytopes (\cref{fig: map3}). The latter environment being larger, we increase the mission time and the depth of the search tree in POMCP from $T_{max}=30$ (Map 1 and Map 2) to $T_{max}=60$ (Map 3).

We do not notice an increase in the time required to update the robot's belief with an increase in problem complexity, which is consistent with our observation that the complexity of the Bayesian update should not increase with the number of obstacles or polytopes. 
On the contrary, the belief update time on Map 2 and Map 3, containing more obstacles, is reduced compared to the first map. More obstacles result in fewer iterations when solving the constrained shortest path problem with $A^{\star}$. Adding constraints due to the obstacles and polytopes reduces the size of the $A^{\star}$ search tree.

\subsection{Limitations}

\textbf{Simulation environments.} In our simulations, we hard-coded the preference policy over the maps (e.g. in Map 1, go around the table counter-clockwise). We randomly sampled problem instances (start and goal locations, and goal options) to reduce the bias introduced by these preference choices. To best evaluate and compare the different approaches, it would be best to sample preferences among a distribution of preferences chosen by a human (for example, from benchmarks resulting from a collection of data). Creating such a benchmark is an interesting direction for future work.

\textbf{Hyperplane arrangement construction.} The main limitation of our approach is that the size and geometry of each polytope depends strongly on the geometry of the obstacles, as seen in \cref{fig:HA}. Because of this, the robot can make predictions over preferences that are too refined compared with the topology of the environment. A direct consequence is that when the size of the polytopes is small, the information provided by the human can be incorrectly interpreted as a preference on the robot's immediate action.
Our method can be improved by changing the structure of the hyperplane arrangement so that it relies on the topology of the environment, but does not vary strongly with the geometry of the features in the environment. For this purpose, topometric maps \cite{badino2011visual} and region construction algorithms \cite{deits2015computing} are promising directions.

% \subsection{Performance Metrics}

% \subsubsection{Inference}
% \begin{itemize}
%     \item Probability of the correct goal wrt time
%     \item entropy of belief distribution wrt time
% \end{itemize}

% \subsection{Baselines}

% Discussion:
% sets strongly depend on geometry of obstacles (use topometric maps as an interesting future direction)

\section{Conclusion}
We presented an approach for encoding and inferring a human's path preference in an environment with obstacles. By leveraging a partitioning of the space into polytopes and a stochastic observation model, our method allows for joint inference over the goal and path preference even when both are unknown a-priori. Our experiments on an unknown-goal navigation problem with sparse human interventions demonstrate the effectiveness of our approach and its suitability for online applications. The time required to update the robot's belief does not increase with the complexity of the environment, which further highlights the practicality of our method. 

%We presented a solution to encode a human's path preference in an environment composed of obstacles, and proposed a problem model leveraging conditional independence assumptions (enabled through our encoding) such that the robot can effectively infer the human's indications through observations in real-time. 

% Future Work

\printbibliography

\end{document}